# Neural Population Decoding and Imbalanced Multi-Omic Datasets For Cancer Subtype Diagnosis

Charles Theodore Kent[1], Leila Bagheriye[2] and Johan Kwisthout[2]
[1]*School of Artificial Intelligence, Radboud Universiteit, Houtlaan 4, Nijmegen, Netherlands*
[2]*Donders Institute for Brain, Cognition & Behaviour, Radboud Universiteit, Houtlaan 4, Nijmegen, Netherlands*
theo.kent@ru.nl, {leila.bagheriye, johan.kwisthout}@donders.ru.nl

Abstract: Recent strides in the field of neural computation has seen the adoption of Winner-Take-All (WTA) circuits to facilitate the unification of hierarchical Bayesian inference and spiking neural networks as a neurobiologically plausible model of information processing. Current research commonly validates the performance of these networks via classification tasks, particularly of the MNIST dataset. However, researchers have not yet reached consensus about how best to translate the stochastic responses from these networks into discrete decisions, a process known as population decoding. Despite being an often underexamined part of SNNs, in this work we show that population decoding has a significanct impact on the classification performance of WTA networks. For this purpose, we apply a WTA network to the problem of cancer subtype diagnosis from multi-omic data, using datasets from The Cancer Genome Atlas (TCGA). In doing so we utilise a novel implementation of gene similarity networks, a feature encoding technique based on Kohoen's self-organising map algorithm. We further show that the impact of selecting certain population decoding methods is amplified when facing imbalanced datasets.

## 1 INTRODUCTION

Multi-omics data integration in cancer diagnosis refers to the integration of information from various biological "omics" e.g., genomics, transcriptomics, metabolomics, to provide a more comprehensive understanding of the molecular landscape of cancer. Spiking neural networks (SNNs) are a neurobiologically inspired method of information processing which aim to solve tasks using plausible models of neuron dynamics (Yamazaki et al., 2022). Much like in biological brains, neurons in SNNs are linked through excitatory and inhibitory connections, and propagate information via discrete electrical signals known as spikes (Yamazaki et al., 2022; Himst et al., 2023). An important feature of SNNs is that their activations are stochastic (Ma & Pouget, 2009), and so presenting a network with the same stimulus multiple times will likely result in varying responses. We can gain more insight into the network through sampling the distribution of responses when presenting a stimulus over multiple time steps, simulating exposure for a given length of 'biological time' (Guo et al., 2017). The responses of the network during this window can be quantified by counting the number of times each neuron spikes, referred to as a spike count code (Grün & Rotter, 2010). Alternatively, some research focuses on the time-dependent relationship of spiking neurons, for instance by weighting neuron responses more highly based on how quickly they fire (Grün & Rotter, 2010; Shamir, 2009; Beck et al., 2008).

In order to extract information from SNNs, we examine the spikes generated by a population of neurons in response to a stimulus. The process of presenting a stimulus to the network to generate these spikes is known as population encoding, and conversely the process of obtaining estimates from the neuron activity patterns is known as population decoding (Ma & Pouget, 2009). Together, these two opposite processes are referred to as population coding. Population coding can be used in conjunction with SNNs to gain practical insights into how a system of spiking neurons tackles the task of learning (Ma & Pouget, 2009).

Approaching the question from a more theoretical standpoint, Bayesian inference is hypothesised to be a key component of information processing within the brain (Guo et al., 2017), including in areas of cognition and decision making (Shamir, 2009). Handling uncertainty when understanding their environment is critical to the survival of many organisms, and Bayes' theorem provides a biologically plausible framework for the brain's probabilistic nature (Kersten et al., 2004). Based on electrophysiological recordings, neurons appear to process information in a hierarchical manner, which can similarly be modelled as hierarchical Bayesian inference (Lee & Mumford, 2003).

Until recently, research into SNNs and hierarchical Bayesian models of the brain have remained separated by the computational complexity of performing exact inference (Guo et al., 2017). To overcome this problem, the variational principle can be employed to decompose the difficult exact inference into an optimisation problem which is easier to solve (Guo et al., 2017; Friston, 2010). Following this approach, spiking neural networks have been able to implement hierarchical Bayesian models through use of neural circuits such as the Winner-Take-All (WTA) circuit (Guo et al., 2017; Nessler et al., 2013). In a WTA circuit, a layer of excitatory neurons is linked to a corresponding layer of inhibitory neurons. Whenever an excitatory neuron fires, an inhibitory signal is generated in response which resets the neuron membrane potentials to baseline and updating the weights of each connection (Himst et al., 2023). When coupled with the spike-timing dependent plasticity (STDP) learning rule, this framework enables neurons to learn structural representations from stimuli in an unsupervised manner (Himst et al., 2023; Guo et al., 2017; Nessler et al., 2013).

Utilising this technique, experimental research into hierarchical Bayesian WTA networks have begun reporting results on the benchmark dataset MNIST (Himst et al., 2023; Guo et al., 2017; Diehl & Cook, 2015; Nessler et al., 2013; Querlioz et al., 2013). Generally, the accuracy of these models on the MNIST test set is in the range of 80-85%, with some works (Guo et al., 2017; Diehl & Cook, 2015) achieving accuracies as high as 95% with an optimised set of model hyperparameters. Whilst the reported results are promising and show the potential applications of WTA networks for real-world problems, current research shows little consideration to the significance of population coding in classification tasks.

One point of contention revolves around the choice of population decoding method used to turn neuronal responses into a discrete prediction for classification. In the vast majority of experiments (Himst et al., 2023; Guo et al., 2017; Diehl & Cook, 2015; Querlioz et al., 2013), neurons are assigned to the class for which they spike most frequently over a given dataset in an a posteriori fashion. Typically, the responses used to make this assignment are collected by presenting the network samples from a training set (Diehl & Cook, 2015; Nessler et al., 2013; Querlioz et al., 2013), however in some cases (Himst et al., 2023; Guo et al., 2017) this step is performed over the test set instead. Unfortunately, the combination of a posteriori assignment and utilisation of test set labels can be shown to lead to high degrees of bias, which we elaborate on in Section 4.

Beyond data subset selection, there are discrepancies between population decoding practices adopted by researchers. By far the most common approach to population decoding (Himst et al., 2023; Guo et al., 2017; Diehl & Cook, 2015; Nessler et al., 2013; Querlioz et al., 2013; Ma & Pouget, 2009) is to assign each neuron a single label based on the class of stimulus for which the neuron responds most highly. Then, upon presentation of a test stimulus, the responses of each neuron are averaged per class, before selecting the class with the highest average firing rate. We term this methodology the *class averaging* decoder, and give a full mathematical description in Section 2.

In Nessler et al. (2013), the population decoding step is hand-performed by a human supervisor by examining the weights of the trained model. Whilst this approach is somewhat reasonable in the context of MNIST, where it is relatively simple for a human to discern the correct classification by eye, it clearly leaves a lot to be desired. Firstly, the process is not scalable, as some notable experimental results (Guo et al., 2017; Diehl & Cook, 2015) recommend using many thousands of output neurons for optimal classification performance. Moreover, not all neurons in the output population will learn a representation that is easily recognisable. A given neuron may be tuned to detect certain sub-features within the image, be half-way between two distinct classes, or fail to learn a meaningful representation entirely and have weights resembling random noise or arbitrary blobs (Himst et al., 2023; Nessler et al., 2013; Querlioz et al., 2013). In these cases, human bias can easily creep into the prediction process, and so a more mathematically grounded approach is desirable.

Querlioz et al. (2013) use a validation subset of 1,000 "well-identified" images to form their neuron-class associations. Querlioz et al. (2013) also point out that the labelling process need not occur concurrently with training, but can be done at a later stage. Furthermore, Querlioz et al. (2013) suggests an avenue for future work could be the coupling of an SNN to a supervised network to perform the population decoding step, a concept that we will be investigating further in this paper.

Notably, in all of the prior discussed approaches, the population decoding step is treated as separate from the SNN model. However, it could be argued that this step must occur somewhere within the brain, as we are ultimately able to resolve sources of uncertainty down into concrete choices. The meta-task of mapping responses from an arbitrarily large population of neurons down to a single discrete decision is generally not approached from a biologically plausible perspective (Ma & Pouget, 2009). As discussed, most methodologies (Himst et al., 2023; Guo et al., 2017; Diehl & Cook, 2015; Querlioz et al., 2013) use a running total of the neuronal responses over each stimulus presented to the network to determine each neuron's class. Yet, it seems implausible for brains to store and update a counter of every time they have seen a certain class of stimuli throughout their whole lives, and reference that counter to make decisions.

A possible alternative to this methodology could be to incorporate a supervised model to perform the population decoding step. For instance, a multivariate logistic regression model requires only the use of a weight, bias and sigmoid activation function; components which have each independently been shown to be neurobiologically plausible (Hao et al., 2020). Another benefit of logistic regression is that the model can be trained in an online fashion, updating the weights and bias upon presentation of each individual stimulus, and thereby avoiding the necessity of viewing the entire dataset simultaneously. Whilst this supervised approach is strictly not biologically plausible, as the learning and inference steps are not spike-based (Hao et al., 2020), these factors at least bring us closer to the desired goal of a complete model of neural information processing.

One of the key factors to be considered in the practical application of Bayesian WTA networks is the role of class imbalance. We posit that the approaches to population decoding which we have discussed play a sizeable role in the system's overall ability to perform classification, and that class imbalance has a strong impact on said performance. Contemporary research primarily focuses on the MNIST dataset, which has equally balanced class samples by default, and so issues which arise in this area have yet to be elucidated. If research is to move beyond the benchmark domain, handling class imbalance is a necessity, as innumerable real-world problems possess this property.

The purpose of this research is to apply an SNN-based hierarchical Bayesian WTA network to a non-benchmark dataset, in order to gain further insight into the implications of selecting various population decoding methods. The remainder of this paper is structured as follows. In Section 2, we introduce the theoretical foundations of population coding in the context of spiking neural networks, as well as the definitions for the population decoding strategies we will test experimentally. In Section 3, the methodology of the experiments is described in detail. Section 4 contains the results of the aforementioned experiments, as well as discussions of the insights gained by practical application of these techniques. Finally, Section 5 contains concluding remarks, suggested areas for further research, and provides our recommendations for future practitioners.

## 2 POPULATION DECODING

In this section, we provide definitions for the methods of population decoding which will undergo experimental evaluation in Section 4. We largely follow the nomenclature provided by Grün & Rotter (2010), in which they discuss resolving the ambiguity of single-trial neuronal responses via population coding.

Consider an experiment in which a spiking neural network is presented with a stimulus $s$ from a stimulus set $S$. Each stimulus has one associated numeric class label $y \in \{0,...,C\}$, where $C$ is the number of possible classes in $y$, and $s_y$ is the classification for a given stimulus. The spikes generated by a population of $N$ neurons in response to presenting a stimulus for a fixed window of time is recorded. The neural population response in this time period is quantified as a vector $r = r_1,...,r_N$ with dimensionality $N$, where $r_n$ is the response of neuron $n$ on a given trial. In this case we are interested in spike counts, so $r_n$ would therefore be the number of spikes emitted by neuron $n$ during the trial in the response window. With this definition of a neural population response, we can now perform various population decoding methods with the response array $r$ to associate the response from a given stimulus $s$ with a predicted classification label $\hat{y}$.

A common strategy (Himst et al., 2023; Guo et al., 2017; Diehl & Cook, 2015; Nessler et al., 2013; Querlioz et al., 2013; Ma & Pouget, 2009) for population decoding is to first associate each output neuron $n$ with a class label present in $\hat{y}$. This association is created based on the relative strength of neuronal responses when reacting to stimuli of each class within the dataset. For each stimulus $s$ presented to the network, we sum the spike counts $r$ of the

output neurons inside a multi-dimensional array $M \in \mathbb{Z}^{N \times C}$ such that

$$M_{nc} = \sum_{s \in S: s_y = c} r_n \quad (1)$$

where the sums of spike count responses for each neuron $n$ are split by class along the $c$ dimension. Each element $M_{nc}$ thus corresponds to the total amount of times a given neuron spiked for a given class over the entire stimulus set $S$.

For each neuron, we can then identify the class which has the highest spike count over the stimulus set. In this way, we can experimentally determine a neuron's preferred class. We represent this associative relationship using the vector $Z = Z_1,...,Z_N$, with dimensionality $N$, defined as:

$$Z_n = \underset{C}{argmax}(M_{nc}) \forall n \in \{1,...,N\} \quad (2)$$

such that each element $Z_n$ represents the preferred class of the corresponding neuron $n$. For ease of notation, we can further treat the vector $Z$ akin to a function, which accepts a parameter $n \in \{1,..., N\}$ representing the index of a neuron as input, and returns the preferred class of the neuron at that index. We denote the neuron's preferred class as $\hat{y}_n$.

$$Z(n) = \hat{y}_n \quad (3)$$

From the definition presented in equations 1 & 2, we can already see there is an implicit assumption that the stimulus set used to construct $Z$ contains a balanced number of examples for each class. This is because the sum of the spike counts is directly proportional to the amount of times a stimulus of that class is presented to the network. In datasets with a high degree of class imbalance, this leads to undersirable behaviour. For instance, a given neuron may have a far stronger response to stimuli of one class relative to another - yet if presented with an overwhelming number of examples of the "less-prefferred" class, the sum of spikes for the less-prefferred class will eventually exceed that of the class with the higher relative spike response rate. This leads to situations where a neuron will be assigned a label in Z which is counter to its observed experimental behaviour. Additional steps must therefore be taken to rectify this behaviour if we wish for SNNs to be performant in class-imbalanced domains.

In the following subsections, we detail the specific population decoding methods being evaluated in this research. Additionally, we make note of each method's potential robustness to class imbalance as a natural result of their mathematical construction.

## 2.1 Winner-Take-All Decoder

For this straightforward population decoding approach, we designate the neuron with the highest spike count response the as 'winner', then find its corresponding preferred class in $Z$ to make the final prediction.

$$Z(argmax(r)) = \hat{y} \quad (4)$$

Overall, the simplicity of this methodology does have significant drawbacks, as each trial is highly sensitive to variability, and the information from the responses of all neurons but the most active is discarded (Ma & Pouget, 2009). In principle, it also has little resistance to class imbalance, as an unequal ratio of neuron labels in $Z$ would cause a disproportionate increase in the likelihood of the majority class being selected.

## 2.2 Population Vector Decoder

Another method is to take a sum of the responses per class, then take the class with the highest amount of 'votes' as the network's prediction (Ma & Pouget, 2009). We split the responses by class in accordance with the observed preferred class of each neuron $Z_n$, such that:

$$\underset{C}{argmax}(\sum_{n:Z_n=c} r_n) = \hat{y} \quad (5)$$

This approach is equivalent to the weighted average shown in (Ma & Pouget, 2009), or is sometimes referred to as 'pooling' the responses of the neuronal population (Grün & Rotter, 2010). This is also the approach implemented in Himst et al. (2023) to achieve their results on the MNIST dataset. Unfortunately, the population vector decoder is greatly susceptible to class imbalance, as it is only concerned with the class-wise sums of responses – thus incurring the imbalance related problems which have been discussed above in regards to construction of the assignment vector $Z$.

## 2.3 Class Averaging

In this approach, we take the highest average firing rate of the neurons per class to determine the prediction. The sum of spike counts for neurons of each class is divided by the number of neurons assigned to that class. Formally,

$$\underset{C}{\operatorname{argmax}}(\frac{1}{Z_c}\sum_{n:Z_n=c} r_n) = \hat{y} \qquad (6)$$

where $Z_c$ is the number of neurons assigned to class $c$ in the preffered class vector $Z$.

This is the methodology adopted by Guo et al. (2017) and Diehl & Cook (2015), and has seen strong experimental results when applied to the MNIST dataset. An interesting mechanism at play in this approach is that, in practice, the distribution of neurons assigned to each class is proportional to the class ratio of the stimulus dataset; a property which we investigate further in our experimental results Section 4. Due to this property, the class averaging decoder is inherently more robust to class imbalance than either of the prior discussed methods.

## 2.4 Firing Average

Here, we propose a novel method of population decoding based upon the average firing rate of each neuron. By subtracting the average firing rate from the spike counts in the response vector, we can thereby pay particular attention to neurons which are abnormally highly active compared to their typical behaviour. We first compute the vector $F = F_1,...,F_N$, pertaining to the average firing rate of each neuron over the a stimulus set of training data:

$$\frac{1}{|S|}\sum_{s \in S} r_n = F_n \qquad (7)$$

where $|S|$ is the cardinality of the stimulus set $S$. We can subsequently subtract the neuron-wise average to obtain the final class estimate of the network.

$$\underset{C}{\operatorname{argmax}}(\sum_{n:Z_n=c} r_n - F_n) = \hat{y} \qquad (8)$$

This approach is theoretically beneficial in reducing the impact of 'over-active' neurons, which are prone to firing regardless of the class of the presented stimulus – effectively acting as a regularization technique. However, what effect this will have on handling class imbalance is as yet unknown. Computationally speaking, calculating the firing average does require an additional pass over the dataset to calculate the F vector. Also, utilising the average spike counts means the values of the vector $F$ are continuous rather than discrete, which further distances this methodology from biological plausibility.

## 2.5 Logistic Regression

As suggested by Querlioz et al. (2013), a viable approach to population decoding could be to couple the SNN to a supervised classification model. To demonstrate this, we consider a multivariate logistic regression model to map network responses to predictions. A variety of other supervised methods could equally apply here, but as the experimental section of this research focuses on a case with a binary target variable, we consider the choice of logistic regression apt for our purposes. Prediction of the target from the network response using the trained logistic regression model is calculated as follows:

$$\frac{1}{1+e^{-(w^T r + b)}} = \hat{y} \qquad (9)$$

where $w$ is the weight vector and $b$ is the bias term. The training procedure is performed in an online manner, updating the weights and bias parameters upon each presentation of a stimulus to the SNN, rather than over the entire dataset at once after the SNN training procedure is completed as with the other population decoding methods.

In regards to the imbalance problem, logistic regression is reasonably adept at handling skewed class ratios. In Section 3.1, we apply a logistic regression model to a heavily imbalanced dataset and observe strong classification performance (shown in Figure 1). This result demonstrates the efficacy of the technique over the original dataset, which suggests it should likewise be able to handle imbalance in the role of a population decoder. Additionally, implementing a supervised model for population decoding negates the necessity of assigning each neuron a discrete class. We therefore do not require the assignment vector $Z$ as in the other described methods, avoiding the implicit problems with class imbalance as discussed prior.

## 3 METHODOLOGY

The dataset we have chosen for practically applying hierarchical Bayesian WTA networks is from The Cancer Genome Atlas (TCGA) (Weinstein et al., 2013). We select the datasets concerning the diagnosis of breast cancer (BRCA) and kidney renal clear cell carcinoma (KIRC). The dataset is comprised of multi-omic features relating to individual patients, including genomics, methylation and mitochondrial RNA sequences. Each patient has a corresponding binary target variable which indicates their cancer diagnosis status, either positive or negative. Importantly for this research, there are far fewer examples of positive diagnoses in both datasets as compared to negative examples, allowing us to investigate the impact of class imbalance. Furthermore, the real-world implications of a false positive versus false negative diagnosis are worth considering. Patients who receive a false positive will likely undergo further tests and ultimately rule out the disease, whereas a false negative could result in the patient going undiagnosed entirely, which can have serious ramifications for treatment outcomes. Therefore, close attention is paid to class-wise predictive performance throughout the methodological process.

### 3.1 Multi-Omic Data

In order to incorporate information from all of the omic types present in the TCGA dataset, the files for methylation, genomics and mitochondrial RNA were combined into a single dataset, with each row representing one patient mapped to approximately 80,000 omic feature columns. We perform separate identical processes for the BRCA and KIRC cancer subtypes. Due to their time-dependent nature, spiking neural networks generally have a high computational complexity. Therefore, it is imperative we perform dimensionality reduction steps upon the dataset in order to maintain a tractable training regime. In this vein, we take after the approach of Fatima & Rueda (2020) and first perform a variance threshold filter over the data. Any feature with a variance of less than 0.2% is removed. This removes any features with zero values recorded for more than 80% of samples, bringing the feature count down to approximately 20,000 for each cancer subtype.

Feature selection is the next step. There are numerous possible algorithms which would be appropriate to apply here; Ang et al. (2015) provides a rich overview of available methods in the specific context of genomic feature selection. As we have labels for our samples, we opt to use supervised feature selection techniques to best make use of all available information in the dataset. In particular, the technique of Minimum Redundancy Maximum Relevancy (mRMR) (Ding & Peng, 2005) has been selected for the purposes of this research. mRMR is concerned with two metrics for feature evaluation; *relevancy* is a measure of the mutual information between a feature and the target, and *redundancy* measures the mutual information between features to select mutually maximally dissimilar genes (Ding & Peng, 2005). These two scores are then considered with equal weight to determine the optimal feature subset.

Using mRMR, we calculate the relevancy and redundancy for each multi-omic feature, and start by selecting the top 20 scoring features. We then perform an ablation analysis upon the selected feature set by training a logistic regression model and sequentially eliminating the lowest scoring remaining feature, noting the degradation in predictive performance each time. In this case we measure performance via F1-score, a decision which is further explained in Section 4. The results of this analysis are presented in Figure 1. Based on these results, we can see that prediction scores reach their maximum by the inclusion of the 10 most relevant features for BRCA and 11 for KIRC. We therefore choose to select the top 11 features for both cancer subtypes, so that the pre-processing phase remains identical in either case.

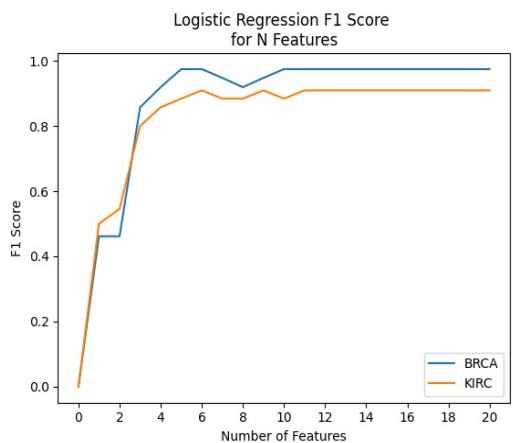

Figure 1: Results of k-folds cross validation for logistic regression models trained on a range of feature subsets. Each time the number of features increases, the new feature being added is the next most important by mRMR score.

### 3.2 Self-Organising Maps & Gene Similarity Networks

Bayesian WTA networks of similar design to that described in this research originate from modelling the processing systems in the visual cortex. The design of Bayesian WTA networks typically incorporates sampling from a Poisson distribution over a given frame of biological time, as a representation of the spiking activity of upstream neuronal firing (Himst et al., 2023; Guo et al., 2017; Nessler et al., 2013). Therefore, an effective way to encode information for processing in Bayesian WTA networks is as a series of binarized images - a 2D grid where each pixel takes the value of either 0 or 1, varying across a time dimension. To accomplish this, the binary images are subsequently encoded into spike trains (Yamazaki et al., 2022), the process of which is further described below in Subsection 3.3.

The chosen method for encoding the selected multi-omic features into an image format is a Self-Organising Map (SOM) (Kohonen, 1990). This technique has been applied to TCGA cancer subtype diagnosis datasets in Fatima & Rueda (2020) with marked success, and so has been selected for the purposes of this research. A further aspect of relevance for this technique is that it has been posited as a biologically-plausible model of neuron self-organisation (Kohonen, 1990). As biological plausibility is likewise a key concern of both hierarchical Bayesian networks and SNNs in general, utilising this technique to encode our input data before presenting it to the network means we can extend this property to encompass the preprocessing stage as well.

We employ Kohoen's Self-Organising Map algorithm (Kohonen, 1990) to translate each feature to a node 2D space, where the Euclidean spatial relationship between nodes encode semantic information about the input data. The organisation is done in an unsupervised manner by iteratively computing the "best matching cell" (Kohonen, 1990) in accordance with the distance between nodes within topological neighbourhoods. Each node has an associated weight vector which is updated concurrently with its local subset, mimicking lateral feedback connections in biophysical network models. The algorithm will run for a set number of epochs or until a desired convergence threshold is reached. Upon completion, the trained SOM returns positional coordinates for each feature in the dataset.

With our newly created spatial feature mappings, we must now generate images representing the omic information of each patient, known as a 'Gene Similarity Network' (GSN) (Fatima & Rueda, 2020). In Fatima & Rueda (2020), samples are encoded via an RGB colour scheme, with each colour channel relating to one of the three types of multi-omic data available in the TCGA datasets. However, since our Bayesian WTA network requires spike trains generated from binarized images as input, using colour to encode information is impossible in this case.

Therefore, we propose a novel method of encoding information into the GSN by scaling and rotating each node in accordance with the strength of feature expression. Each feature is first normalised by Z-score to reduce the impact of outliers. Then, to determine the size of the GSN nodes, each feature is scaled between a range of minimum to maximum desired pixel sizes, proportional to the overall size of the generated image. To determine the orientation of each node, we similarly scale the feature columns between the range of 0 and 180 degrees, as we opted to use a diamond shape for each node with an order of rotational symmetry of 2. We run the SOM algorithm on our dataset for 5 epochs with a learning rate of 0.05. An example of the completed GSNs is shown in Figure 2.

Encoding information via orientation for WTA networks is a well-studied approach given the origins of this research focus on processing in the visual cortex of various animal species (Grün & Rotter, 2010; Ma & Pouget, 2009). On the whole however, binary images are a somewhat limiting format for encoding information, as there is only one degree of granularity for each pixel feature. This makes the task of encoding continuous data into binary pixels a challenging one, and we identify this as an area for potential future research. The GSN implementation presented here attempts to overcome this information bottleneck by using conjointly utilising location, rotation, and size of shapes within the image.

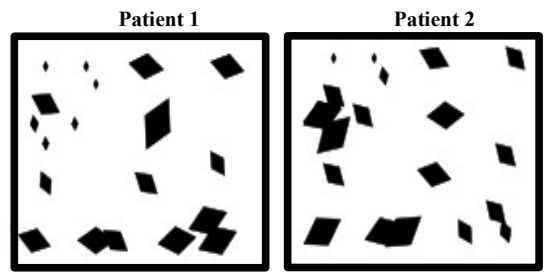

Figure 2: Gene Similarity Network for two patients. Each diamond represents one feature and shares positions across patients, but varies in size and orientation based on each patient's level of feature expression.

## 3.3 Spike Trains

The time-dependent nature of SNNs requires that stimuli be presented to the network over an extended period of time, so as to model biological processing (Guo et al., 2017). However, research has shown (Guo et al., 2021) that presenting one static input for the duration of the presentation is both inefficient for learning and questionable in terms of biological plausibility. Instead, it is preferable that the stimulus has variability over time. To accomplish this, we follow the procedure of (Himst et al., 2023; Guo et al., 2017; Nessler et al., 2013). The binary GSN images are converted into Poisson spike trains, where pixel values for each timestep are drawn from a Poisson distribution modulated by the colour (white or black) of that pixel in the original image. We select a firing rate of 200hz for generating the spike trains, and present them to the WTA network for 150ms of simulated biological time.

### 3.4 Synthetic Minority Oversampling

As certain methods of population coding are potentially highly sensitive to class imbalance, one particularly useful tool in this circumstance is that of Synthetic Minority Oversampling Techniques (SMOTE) (Chawla et al., 2002). SMOTE offers an effective way to mitigate imbalance-related issues by including additional synthetic examples of the minority class in the training set. Although there are numerous potential methods to generate new synthetic datapoints, for the purposes of this research we deem it sufficient to simply over-sample the minority class up to a desired ratio of class imbalance. This is due to the fact that several of the population decoding methodologies described in Section 2 are heavily affected by class imbalance; in these cases, the impact of training on more varied samples has a negligible impact on predictions as compared to merely re-balancing the training class distribution. We define the class ratio $\alpha$ of a set as:

$$\alpha = C_m / C_M \quad (10)$$

where $C_m$ is the number of samples in the minority class, and $C_M$ is the number of samples in the majority class (Imbalanced-learn, 2016). Prior to resampling, the BRCA dataset has a ratio of $\alpha=0.066$ and KIRC has a ratio of $\alpha=0.091$. We investigate the impact of various $\alpha$ ratios experimentally in Section 4.

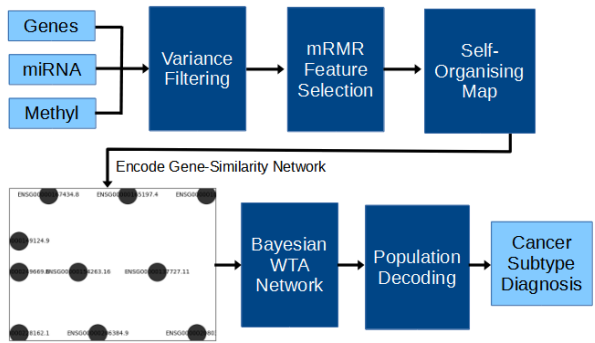

Figure 3: Diagram representing the methodological process for this research. We start with multi-omic data, apply pre-processing steps, and encode into a binary image. These are used to train a Bayesian WTA network, the responses from which we can then use various methods to decode into a final prediction from the system.

## 4 EXPERIMENTAL RESULTS

In this section, we experimentally evaluate the performance of a hierarchical Bayesian WTA network on the TCGA datasets for the BRCA and KIRC cancer subtypes. The network we choose for our experimentation is based on the design presented in Guo et al. (2017), making use of the code implementation provided by Himst et al. (2023). The network is composed of an input, hidden, and output layer. The shape of the input layer is determined by the pixel size of GSN images, which is 176 x 128. We split the image into 16 subsections of size 11 x 8. Each of these 16 sensory blocks then feeds into a layer of WTA circuits with 32 hidden neurons. Finally, each of the neurons in the hidden layer is connected through a single WTA circuit consisting of 100 output neurons. The network includes top-down connections as suggested by Himst et al. (2023) in an effort to improve the network's learning and classification performance. To evaluate the classification performance of our methodologies, we use the metric of F1 score. F1 score was chosen over the typical accuracy metric for classification, as the TCGA datasets contain heavy class imbalance. Due to the nature of many population coding methods, it becomes trivial to achieve a high accuracy score by training a network which only predicts the majority class regardless of the input. In fact, this is an outcome which we must take steps to actively avoid in some cases, such as by applying SMOTE to the dataset. Furthermore, F1 score gives a higher weighting to the classification performance of the positive class, which is pertinent for cancer detection due to the asymmetrical real-life ramifications of reporting a False Negative versus False Positive result. In the case of all experiments involving SMOTE, the F1 score is reported only on data present in the original dataset.

### 4.1 Effects of Imbalanced Datasets on Population Coding

One of the key goals of this research is to highlight how different methods of population coding respond to class imbalance. In order to demonstrate this, we use SMOTE to adjust the class ratio $\alpha$ of the training sets for the WTA network, and perform K-folds cross validation on the rebalanced datasets, retraining the network each time. We can then test each of the population decoding methodologies described in

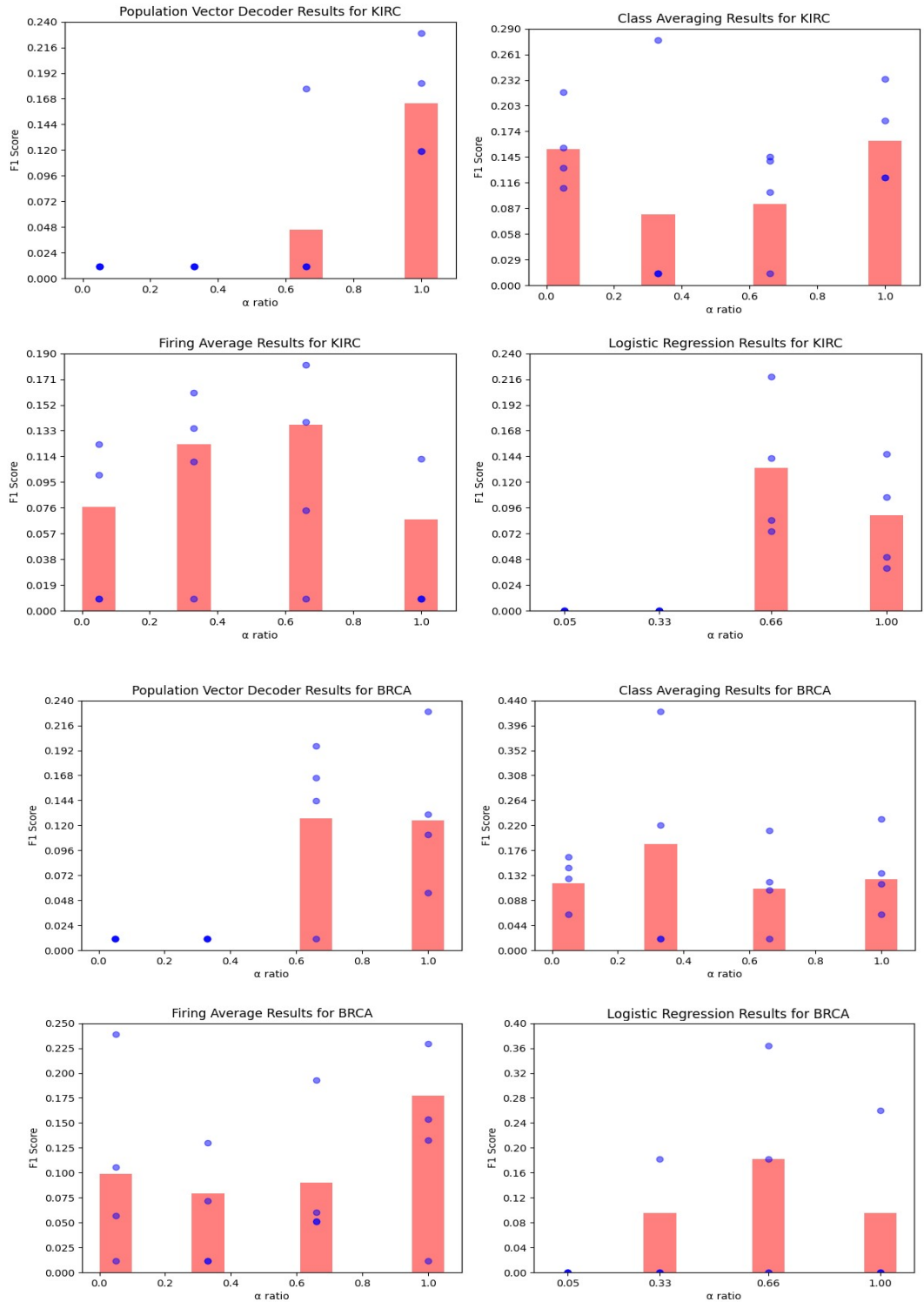

Figure 4: Results for different population decoding methods of SNNs trained on datasets with varying levels of synthetic oversampling. Each red bar represents the score over the entire dataset, whereas each blue dot represents the score for a single fold.

Section 2. The results of this experiment are shown in Figure 4. As we can see from Figure 4, the performance of the population vector decoder has a strong positive correlation with **α**. Not pictured in Figure 4 is the winner-take-all decoder, which had a consistent F1 score of zero both on each training fold and over the entire dataset, regardless of **α**. In the case of where no SMOTE was applied and the network is trained on the standard TCGA dataset, both of these methods report an F1 score of zero. Probing deeper, this is due to every neuron in the output population being associated with the majority class, thereby rendering the system unable to make predictions of the minority class. Logistic regression had similar issues with performance on the default dataset with a low **α** ratio, but saw a marked improvement at the point of oversampling to **α**=0.33 for BRCA and **α**=0.66 for KIRC, and performed reasonably well above these thresholds. Class averaging and firing averaging both performed considerably better across all **α** values. They therefore demonstrate resilience to class imbalance, as there appears to be no strong correlation between their predictive performance and **α** ratio. Across both datasets, class averaging had the highest single performance of any population decoding method trialled in this research.

## 4.2 Distribution of Neuron Class Assignments

Herein lies an exploration of neuron class assignments. Presented in Figure 5 is a heatmap of the **α** ratio of neuron class assignments determined via Equation (2). We further calculate the Pearson correlation coefficient between the **α** ratio of the training dataset and the neuron class assignments: The BRCA dataset has a dataset-neuron **α** correlation coefficient of 0.932, and KIRC 0.879. Both datasets showing such a strong correlation is certainly indicative of the relationship between the distribution of neuron classes and classes in the training set. This result is notable as one may expect, for instance, that the number of neurons assigned to a certain class be dependent upon the complexity of the stimuli within that class. Querlioz et al. (2013) ascribe the improvement of predictive performance when increasing the size of the population of output neurons to the notion that the population is able to learn more diverse representations of the output class. However, as our chosen SMOTE technique is to simply oversample the minority class, complexity of the stimuli is constant across varying levels of **α** ratio. From these results, we suggest that the determining factor for the class assignment would therefore appear to be the class distribution of the stimulus set. Another point of interest is that whilst the correlation between the **α** ratios is high, the heatmap in Figure 5 shows that the relationship is not exactly linear. On the unmodified datasets, the low **α** ratio leads to mode collapse, where every neuron in the output population is assigned to the majority class. For the highest **α** ratio, where the dataset was synthetically oversampled to have an equal class distribution, the neuron class distribution also reaches a similarly high **α** ratio. However, for both the **α**=0.33 and **α**=0.66 datasets, there is a much larger discrepancy between the dataset and neuron assignments. This is likely indicative of the shortcomings of Equations (1) & (2) when dealing with class imbalance which we introduced in Section 2 – regardless of whether a neuron is presented with 3 or 6 examples from the minority class, if it's shown 10 from the majority class then it has a considerably greater likelihood of being assigned as a majority neuron. This hypothesis further explains the "jump" in neuron assignments when moving from **α**=0.66 to **α**=1.0, as the minority class is finally placed on equal footing with the majority class.

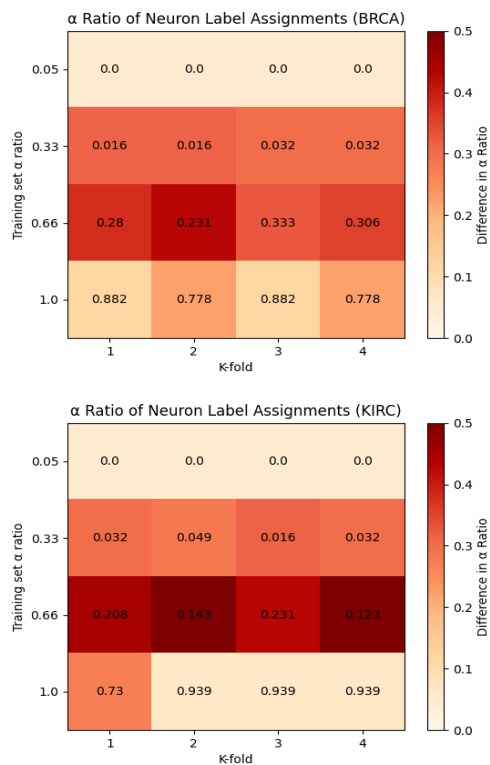

Figure 5: Heatmap of the relationship between the **α** ratio of neuron class assignments versus the class distribution of the training set. The numerical value within each cell is the **α** ratio of neuron assignments for each of the K-folds during training. The colour scale represents the absolute difference between the neuron assignment **α** ratio and the **α** ratio of the training set.

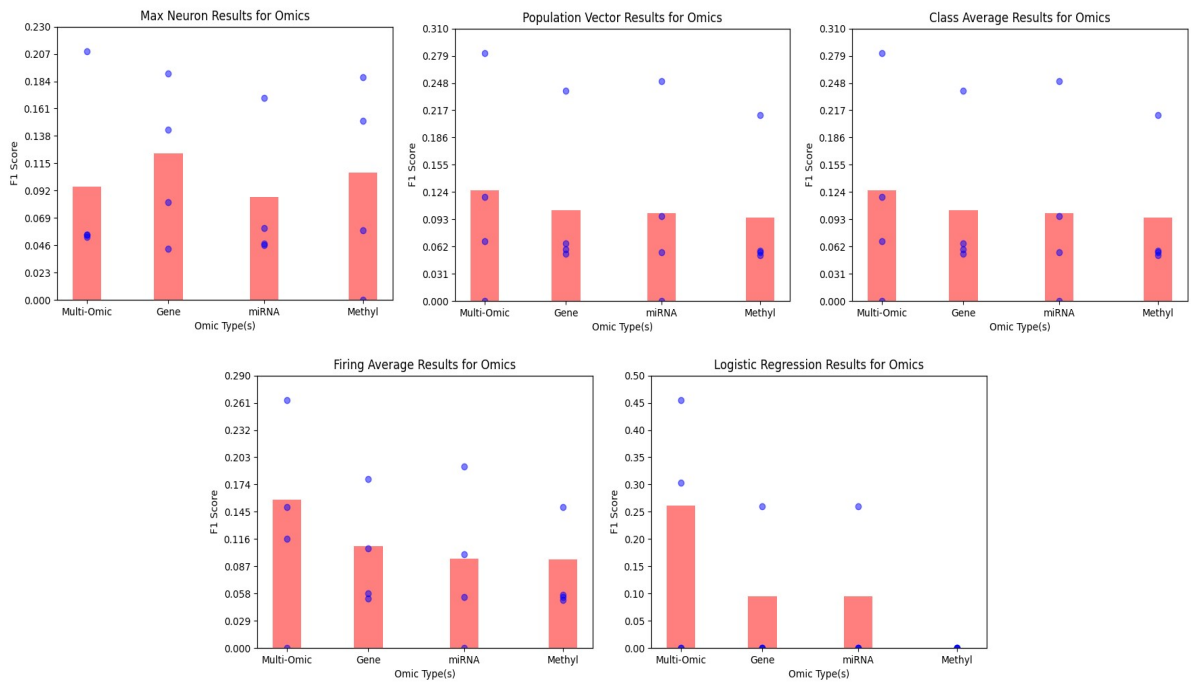

Figure 6: F1 score results for the WTA network trained on various subsets of omic information using K-folds cross validation. Each bar represents the score over the entire dataset, whereas each blue dot represents the score for a single fold.

### 4.3 Multi versus Single Omics

In this section, we analyse the network's performance when trained on subsets of the omic information present in the BRCA dataset. The results of these experiments are shown in Figure 6. Our results generally concur with that of other researchers in relation to the predictive power of the omic types (Fatima & Rueda, 2020). Utilising all multi-omic features together leads to the best classification results. This is followed by genomic, mitochondrial RNA, and methylation features (respectively). These results are encouraging as they support the network's claim to be effectively learning information from the input stimuli, despite the fact that the overall classification is quite poor in comparison with other techniques applied to TCGA datasets (Fatima & Rueda, 2020).

### 4.4 Test Set Bias

Here, we demonstrate the effects of the bias introduced by performing population decoding using the responses to a test set, and then predicting on that same test set. First, consider using a trained WTA network to perform classification upon a test set, using any of the population decoding methods defined in Section 2 where $Z$ is a prerequisite (i.e. not logistic regression). In this example, we set the initial size of the test set to only one stimulus sample. Following Equation (1) to construct Z, we notice that it is impossible for the system to produce an incorrect prediction; whichever neuron(s) spiked when presented with the stimulus are retroactively assigned to be a neuron of that class. Shown in Figure 6 is the accuracy score when using a completely untrained network's responses over various sized subsets of the testing set to determine neuron class associations. We use accuracy score in this case rather than F1, as the randomly distributed test subsets may not contain any positive samples, thereby making F1 inapplicable. We can clearly see from Figure 7 that the "infallible neuron" problem arises when the size of the test set has a low number of samples. We can further determine that the bias introduced by this methodology decreases as the number of test samples increases. This is likely why the issue was not identified by Himst et al. (2023) & Guo et al. (2017) when working with the 10,000 sample MNIST testing set, but is certainly something to be wary of with smaller datasets. We posit that this may also be a contributing factor to the ability of these systems to have such high performance whilst only being exposed to the training set for one epoch, whereas

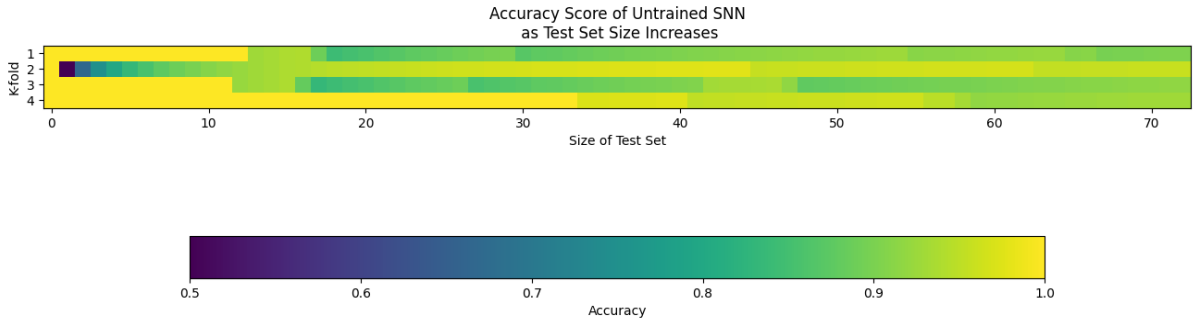

Figure 7: Heatmap of the accuracy score results of an untrained SNN making predictions upon a test set of increasing size, where the population decoding is calculated using the responses of the network to that same test. The results shown here are calculated over the BIRC dataset split into 4 folds.

Nessler et al. (2013), Diehl & Cook (2015) and Querlioz et al. (2013) all required multiple epochs to achieve similar results. Luckily, the solution we propose to address this bias is extremely simple: use the network responses from the training set to perform population decoding, rather than the test set responses. This removes the need to use test labels to make predictions, and importantly eliminates the infallible neuron problem. One additional piece of advice for practitioners making this change is to only start record training responses after the network has been trained for a desired number of epochs, as using the responses from an untrained network is liable to give poor results. Diehl & Cook (2015) show that following these practices can still lead to strong classification performance.

## 5 CONCLUSION

Throughout this work, we have explored the motivation and implications of implementing an array of both common and novel population decoding strategies for multi-omic based cancer subtype diagnosis. Our findings show that the winner-take-all and population vector decoders are both heavily impacted by class imbalance, whereas class averaging, logistic regression, and our novel firing average implementation are more imbalance resistant. We further show that the assignment of neuron classes in population decoders is highly correlated with the class distribution of the stimulus set. This is a property which has, as far as we can identify, gone unnoted thus far in the literature, and warrants further investigation into the relationship between the complexity of stimuli, class distribution and neuron assignments.

Overall, the predictive performance of our system is poor in comparison to other methods using this dataset (Fatima & Rueda, 2020). In our estimation, the primary issue lies in the loss of information when transforming multi-omic data into binarized GSN images. We can observe from Figure 1 that a simple logistic regression model is capable of achieving near-perfect results using the same feature subset, thereby isolating the problem to either the GSN or the SNN. We later tested the classification performance of a simple Convolutional Neural Network (CNN) on the GSN images, which likewise had difficulty extracting information from the binary data, and surprisingly lead to even poorer predictive performance than the SNN system. Thus, we conclude that more sophisticated methods of information encoding are necessary if we wish to apply SNNs to datasets of within the domain of multi-omics, or indeed to continuous tabular datasets in general.

Despite the quality of our predictions, we are nevertheless able to identify issues with current practices in regards to the bias introduced by utilising testing set labels and responses to perform population decoding. Our recommendation for future researchers is that this practice be avoided in favour of using the training set, with the caveat that the network be trained for at least one epoch before collecting the responses.

Several challenges that we faced during this research were related to the highly stochastic nature of Bayesian WTA networks. This leads to high variance in training convergence, making the system's performance difficult to evaluate in general terms. K-folds cross validation is absolutely necessary in this instance to gain insight into the variance of results between runs. Furthermore, due to their non-parallelizable nature and high dimensionality requirements, training times can be exceedingly long (Querlioz et al. (2013) report approximately 8 hours for one run on the MNIST dataset). Combined, these two factors make iterative improvement difficult, as well as making it intractable to explore high dimensional hyperparameter spaces. One area for future research could therefore be the application of more computing power to properly perform hyperparameter optimisation on the network, which could lead to superior performance.

Future research may alternatively wish to focus on a biologically plausible method of translating population responses into discrete decisions. One potential direction for this is suggested by Hao et al. (2020), wherein they combine the unsupervised STDP learning rule with a leaky integrate-and-fire neuron model to perform classification on the MNSIT dataset. We have identified the encoding of information as a bottleneck, as SNNs necessitate the discretisation of information into spikes, inherently impairing the maximum information density of the system. To perhaps alleviate this problem, further research could attempt alternative spike encoding methods, such as those suggested by Guo et al. (2021). On the other end of the system, interesting insights could be gained by investigating the temporal nature of neural responses, as opposed to the spike count code (Grün & Rotter, 2010).